\documentclass[journal]{IEEEtran}

\usepackage{framed,multirow}

\usepackage{amssymb}
\usepackage{latexsym}

\usepackage{url}

\usepackage[utf8]{inputenc} 
\usepackage[T1]{fontenc}    
\usepackage{hyperref}       
\usepackage{url}            
\usepackage{booktabs}       
\usepackage{amsfonts}       
\usepackage{nicefrac}       
\usepackage{microtype}      
\usepackage{graphicx}
\usepackage{chngpage}
\usepackage[ruled,vlined]{algorithm2e}
\usepackage[table, dvipsnames]{xcolor}
\definecolor{lightgray}{gray}{0.75}
\definecolor{lightergray}{gray}{0.95}
\usepackage[multiple]{footmisc}
\usepackage{enumitem}
\usepackage{amsmath}

\DeclareMathOperator*{\argmin}{arg\,min}

\usepackage{authblk}
\usepackage{lipsum}

\begin{document}



\title{Local contrastive loss with pseudo-label based self-training for semi-supervised medical image segmentation}




\author{\IEEEauthorblockN{Krishna Chaitanya, Ertunc Erdil, Neerav Karani, and Ender Konukoglu}
\thanks{KC, EE, NK and EK are with the Computer Vision Laboratory, ETH Zurich, Switzerland. Email: \{krishna.chaitanya\}@vision.ee.ethz.ch. \\
Under review at a Journal.}}

\maketitle

\begin{abstract}
Supervised deep learning-based methods yield accurate results for medical image segmentation. However, they require large labeled datasets for this, and obtaining them is a laborious task that requires clinical expertise.
Semi/self-supervised learning-based approaches address this limitation by exploiting unlabeled data along with limited annotated data.
Recent self-supervised learning methods use contrastive loss to learn good global level representations from unlabeled images and achieve high performance in classification tasks on popular natural image datasets like ImageNet.
In pixel-level prediction tasks such as segmentation, it is crucial to also learn good local level representations along with global representations to achieve better accuracy.
However, the impact of the existing local contrastive loss-based methods remains limited for learning good local representations because similar and dissimilar local regions are defined based on random augmentations and spatial proximity; not based on the semantic label of local regions due to lack of large-scale expert annotations in the semi/self-supervised setting.
In this paper, we propose a local contrastive loss to learn good pixel level features useful for segmentation by exploiting semantic label information obtained from pseudo-labels of unlabeled images alongside limited annotated images.
In particular, we define the proposed loss to encourage similar representations for the pixels that have the same pseudo-label/ label while being dissimilar to the representation of pixels with different pseudo-label/label in the dataset.
We perform pseudo-label based self-training and train the network by jointly optimizing the proposed contrastive loss on both labeled and unlabeled sets and segmentation loss on only the limited labeled set.
We evaluated the proposed approach on three public medical datasets of cardiac and prostate anatomies, and obtain high segmentation performance with a limited labeled set of one or two 3D volumes.
Extensive comparisons with the state-of-the-art semi-supervised and data augmentation methods and concurrent contrastive learning methods demonstrate the substantial improvement achieved by the proposed method.
The code is made publicly available at https://github.com/krishnabits001/pseudo$\_$label$\_$contrastive$\_$training.
\end{abstract}




\section{Introduction}\label{sec:intro}
Medical image segmentation with high accuracy is very desirable for many downstream clinical applications.
Currently, supervised deep learning methods yield state-of-the-art segmentation performance~\cite{ronneberger2015u,milletari2016v,kamnitsas2016deepmedic,kamnitsas2017efficient}, however, this approach requires having access to large labeled datasets.
Obtaining such large labeled datasets for segmentation is time-consuming and expensive in medical imaging since the annotations need to be done by clinical experts.  
To attenuate the need for large labeled datasets, semi-supervised~\cite{lee2013pseudo,rasmus2015semi} and self-supervised learning methods~\cite{doersch2015unsupervised,chen2020simple} leverage unlabeled data along with limited labeled examples.
Recent methods have yielded promising results on medical datasets and reduced the requirement for a large number of the labeled examples~\cite{bai2017semi,chaitanya2020contrastive}.

\subsection{Motivation}
Prominent works in semi-supervised learning methods focus on extracting useful information from unlabeled examples along with using limited labeled examples.
Some popular semi-supervised learning strategies include using pseudo-labels for self-training~\cite{lee2013pseudo,bai2017semi}, entropy minimization~\cite{grandvalet2005semi}, consistency regularization~\cite{sajjadi2016regularization,laine2016temporal} and data augmentation~\cite{kcipmi2019,zhao2019data}. 
Similarly, self-supervised learning methods such as pretext-task-based~\cite{doersch2015unsupervised} and contrastive learning-based~\cite{chen2020simple} methods aim to learn a good model initialization via pre-training with only unlabeled images and later fine-tune this initialization with limited annotations to get high performance.

Contrastive learning-based methods yielded highly accurate models on many natural and medical image datasets for classification and segmentation tasks.
The objective of contrastive learning~\cite{hadsell2006dimensionality} is that the latent representations of similar images should be alike, and simultaneously they should be different from the representations of the dissimilar images.
Therefore, it is crucial to correctly define similar and dissimilar images during optimization to obtain better representations.
Earlier works~\cite{chen2020simple} define similar images as two differently transformed views of an image while defining dissimilar images as images with different contents. 
Some recent methods~\cite{chaitanya2020contrastive} define similarity cues by using domain knowledge for medical images that go beyond random transformations by defining an additional set of similar images as slices from corresponding anatomical areas across subjects.
These methods learn global-level image representations useful for downstream tasks such as classification and segmentation and achieve high performance when fine-tuned with limited labels.

\begin{figure}[!ht]
    \centering
    \includegraphics[width=0.5\textwidth]{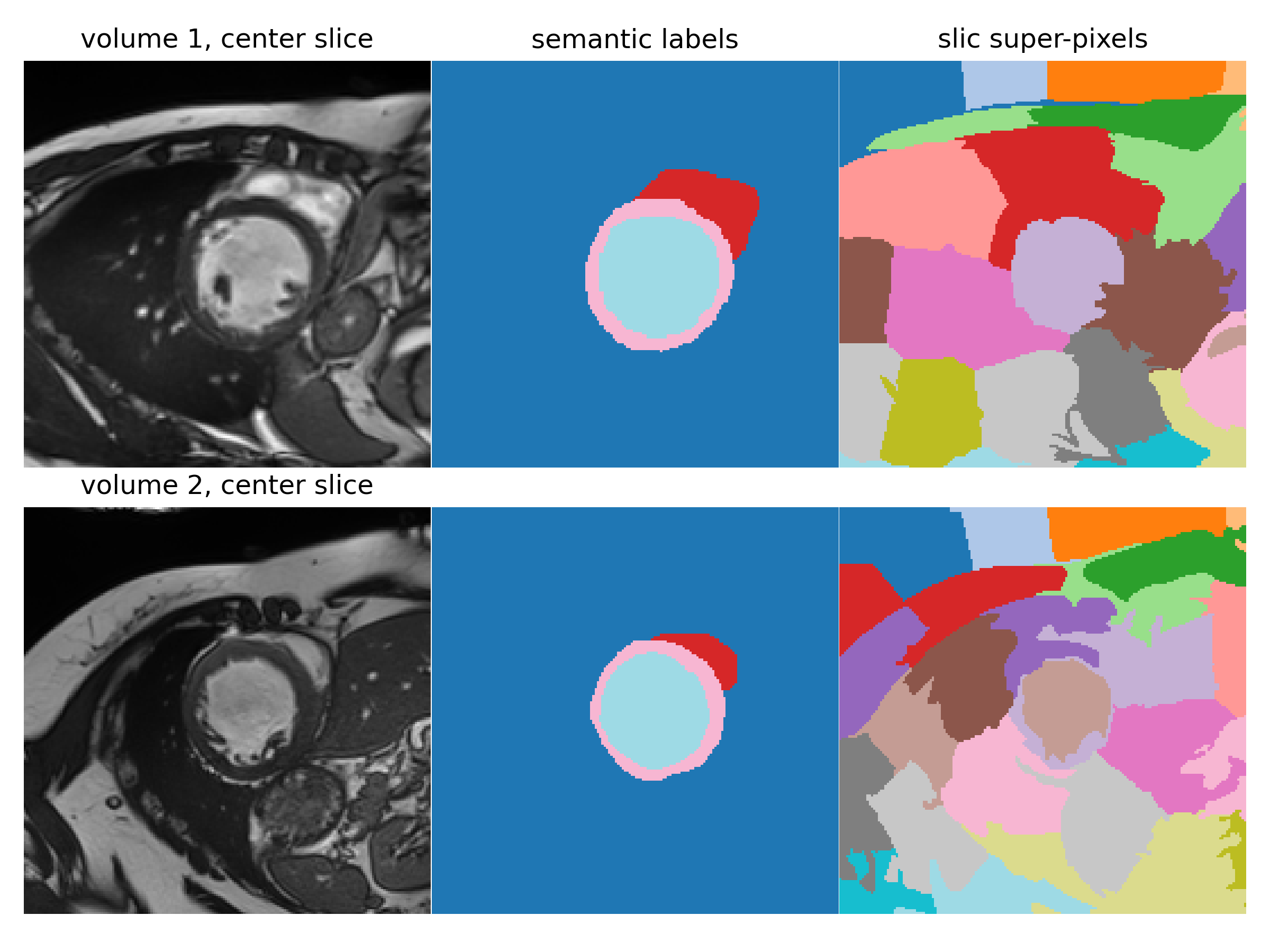}
    \caption{A visual example that demonstrates the shortcoming of existing local contrastive learning methods. Unsupervised segmentation methods such as a super-pixel based one (Slic~\cite{achanta2012slic}) 1) do not produce semantic labels similar to ground truth, 2) do not assign the same label id for the same anatomy from different images (see different colors assigned by Slic for the similar regions) which hinders to enforce similarity/dissimilarity of pixel representations of same anatomies across images~\cite{henaff2021efficient}}
    \label{fig:slic_super_pixel}
\end{figure}

Unsupervised learning of good local image features can be as crucial as global features for pixel-level prediction tasks such as segmentation.
However, the performance of local contrastive learning-based methods is rather limited compared to global ones due to the difficulty of defining similar/dissimilar local regions without semantic labels.
Early work by~\cite{chaitanya2020contrastive} enforce patch-level local features across different augmented views of an image to be similar to each other while being dissimilar to representations of other local regions within the image.
More recent local contrastive learning methods extend this work by using surrogate semantic labels~\cite{henaff2021efficient,van2021unsupervised} where the labels are estimated on unlabeled images using supervised techniques such as saliency maps estimation approaches~\cite{nguyen2019deepusps,pinheiro2015learning,arbelaez2010contour}, super-pixels~\cite{achanta2012slic}, and image computable masks~\cite{felzenszwalb2004efficient,arbelaez2014multiscale}.. 
These methods encourage local regions having the same semantic label to have similar representations while being dissimilar to regions with different labels.
There are two possible shortcomings of using unsupervised methods to obtain surrogate segmentation masks of unlabeled images: 1) surrogate class labels obtained by an unsupervised method most likely do not match with the desired target semantic classes, 2) similarity/dissimilarity of local regions across different images cannot be enforced since these methods do not necessarily assign the same label id for the same anatomy across different images. This is illustrated in Figure~\ref{fig:slic_super_pixel}. These problems may limit the quality of the learned representations. Also, these approaches consist of two stages training: pre-training with surrogate labels and fine-tuning with limited target labels for any downstream tasks like classification or segmentation.
This motivates us to formulate an end-to-end joint training framework suited for segmentation task where we devise to use pseudo-labels of unlabeled data instead of surrogate labels from unsupervised methods to overcome these problems and achieve high segmentation accuracy.

\subsection{Contributions}
(a) In this paper, we address the above limitations by proposing an end-to-end joint training semi-supervised approach more suitable for segmentation task by defining a local pixel-level contrastive loss on pseudo-labels of unlabeled set and limited labeled set.
The proposed contrastive loss encourages similar representations for the pixels that belong to the same target semantic label while simultaneously enforcing them to be dissimilar to the representations of the pixels from different classes.
The labels for the unlabeled set are assigned with the pseudo-label estimates obtained by passing the unlabeled set through an initial network trained with a limited number of labeled images.
The pseudo-labels are updated for every pre-defined number of iterations during the training to get better pseudo-label estimates. The network is jointly trained using the proposed local contrastive loss on unlabeled and labeled sets, and supervised segmentation loss computed only on the limited labeled set in an iterative fashion like self-training~\cite{bai2017semi}. 


(b) The most relevant concurrent works are~\cite{alonso2021semi,zhao2020contrastive}.
In~\cite{zhao2020contrastive}, first they performs pre-training with label-based contrastive loss and later fine-tune with segmentation loss where both steps are fully supervised. Unlike~\cite{zhao2020contrastive}, we perform an end-to-end joint training with both contrastive and segmentation losses and also leverage pseudo-labels only in the contrastive loss.
In~\cite{alonso2021semi}, authors propose an end-to-end training in a student-teacher framework with both contrastive and segmentation losses computed on both pseudo-labels and true labels. They use a memory bank to maintain and update the feature representations of each class, and pseudo-labels are estimated by teacher network, and also perform entropy minimization on unlabeled predictions.
Here, we use a much simpler approach with only one network for training and continuously update the pseudo-labels with re-estimation step, and do not use any external memory and entropy minimization. Also, we explicitly do not use the pseudo-labels in the segmentation loss unlike~\cite{alonso2021semi} which we observed to improve the results further.

(c) The defined local loss objective enforces intra-class similarity and inter-class separability for the target semantic classes across the dataset desirable for the segmentation task, while the previous unsupervised local contrastive loss methods can enforce this only within an image since unsupervised segmentation methods like Slic~\cite{achanta2012slic} do not create consistent semantic label ids across images. 

() We evaluated the benefits of the proposed method on three public MRI medical datasets and compared the performance with state-of-the-art self-supervised and semi-supervised learning methods.
The results demonstrate that the proposed method improves the best performing baselines.

(e) We perform a detailed ablation study to understand how the performance is affected by each component or loss term defined in the learning framework. Also, we evaluate if using consistency loss to use only the highly confident pseudo-labels influences the performance gains obtained.

\section{Related work}
We broadly classify the relevant literature into below categories with respect to the proposed method:

1. Self-supervised learning (SSL)
Many recent works using SSL with appropriate unsupervised loss have shown to learn useful representations from unlabeled data. Models obtained from such optimization serve as good initialization for various downstream tasks and yield performance gains over random initialization. 
SSL works relevant to the presented work can be categorized into below two categories:

(i) Pre-text task-based methods: Here, a pre-text task is devised whose labels can be acquired with ease and freely from the unlabeled data itself to learn the initialization. Some examples of such tasks include: predicting the rotation applied on the input image~\cite{gidaris2018unsupervised}, in-painting the missing pixel values~\cite{pathak2016context}, context restoration~\cite{chen2019self}, and many more~\cite{doersch2015unsupervised,noroozi2016unsupervised,zhang2016colorful,dosovitskiy2014discriminative}.

(ii) Contrastive learning methods: These methods~\cite{wu2018unsupervised,oord2018representation,hjelm2019learning,chen2020simple,he2019momentum,henaff2019data,misra2019self} use a contrastive loss~\cite{hadsell2006dimensionality,gutmann2010noise} to enforce the representations from positive pairs to be similar and simultaneously be dissimilar to other representations. In general, positive pairs are defined by applying two different transformations (random augmentations) on an image. While some recent approaches use the domain-specific knowledge to capture more complex similarity cues over random augmentations by defining additional similar and dissimilar pairs, which have shown improved results for videos~\cite{tschannen2020video} and medical images~\cite{chaitanya2020contrastive} over just using random transformations on an image. Most of these methods are defined to learn good global level representations that are useful for classification tasks.
For dense prediction tasks such as segmentation, there are some relevant works which we list into below three categories:

(a) Supervised local contrastive learning: In these works~\cite{zhao2020contrastive,wang2021exploring,lee2021attention} authors have access to many labeled examples which they use to define an additional contrastive loss over standard segmentation loss to learn discriminative local pixel level features for the similar semantic classes over the whole dataset. One work~\cite{hu2021semi} explores this supervised approach on medical image segmentation with limited annotations and get some improvements.
In this work, we focus on the scenario where only limited labels are available.

(b) Unsupervised local contrastive learning: Here, the authors do not have access to ground-truth masks. Some methods~\cite{chaitanya2020contrastive,xie2020pgl,wang2020dense,jaus2021panoramic} pre-train by matching patch/pixel-level representations across either across two augmented views of an image or positive pair of images defined using domain cues.
Other methods\cite{van2021unsupervised,henaff2021efficient} obtain some surrogate masks of fore-ground objects using unsupervised techniques such as saliency maps~\cite{nguyen2019deepusps,pinheiro2015learning,arbelaez2010contour}, super-pixels~\cite{achanta2012slic}, image computable masks~\cite{felzenszwalb2004efficient,arbelaez2014multiscale}. Later, the local level features of a chosen fore-ground object are optimized to be similar while other objects local features are optimized to be dissimilar. Here, the similar pixels or regions are chosen for the same object across two different transformations applied on an image. Often the fore-ground object classes estimated via the surrogate masks may not resemble with any of the multiple target classes to be segmented in medical images as shown in Fig.~\ref{fig:slic_super_pixel}. Also, these methods like Slic~\cite{achanta2012slic} do not generate consistent label ids for similar structures across different subjects. Therefore, in training, we cannot capture the complex similarity cues available for the same class across different subjects due to lack of true labels.

(c) Semi-supervised local contrastive learning:
Recently, some concurrent works~\cite{alonso2021semi,zhao2020contrastive,peng2021boosting,you2021momentum,xiang2021self,zhou2021c3,you2021simcvd} have devised semi-supervised learning frameworks that use unlabeled images with some variant of contrastive loss setup.
The works relevant to presented work are~\cite{alonso2021semi,zhao2020contrastive}.
In~\cite{alonso2021semi}, they use the pseudo-labels of unlabeled images in both cross-entropy loss and contrastive loss while ~\cite{zhao2020contrastive} use a two step iterative training process: pre-training with contrastive loss and later fine-tuning with both labeled annotations and pseudo-labels of unlabeled set for the downstream task. We d
We only use the pseudo-labels of unlabeled images for contrastive loss computation and avoid using them in the segmentation loss (e.g., cross entropy or Dice loss) computation. Also, we train the network in one step with a joint training setup using both contrastive and cross-entropy losses simultaneously. 
Also, in~\cite{alonso2021semi} they use a student-teacher network, employ entropy minimization on unlabeled set predictions, and use a separate memory bank to store and update the feature representations of the each class used in the training. We propose a much simpler setup with only one network and do not use any memory bank or entropy minimization.
In limited annotation settings, where pseudo-labels are not very accurate, we observe empirically in our experiments that using them directly in cross-entropy loss hinders the network from obtaining higher performance gains. We observe the same for the two step training method. These results are presented in Table~\ref{table:concurrent_cont_lr_works}. 
Similar to~\cite{alonso2021semi}, some other recent/concurrent works~\cite{zhou2021c3,you2021simcvd} also apply such contrastive loss with the teacher and student networks.  
In~\cite{zhou2021c3}, they apply a pixel-wise contrastive loss on highly confident predicted regions from the same class, and additionally leverage consistency loss between predictions of the teacher and student networks.
In~\cite{you2021simcvd}, they apply the contrastive loss alongside segmentation and other consistency losses. Here, the contrastive loss is designed to learn object shape information with the help of boundary-aware representations, defined on the predicted signed distance maps of teacher and student networks.


2. Semi-supervised learning
There has been enormous amount of works in semi-supervised learning that leverage unlabeled and labeled images together for training. Below we only describe three sets of works relevant to the proposed work.

(a) Self-training: 
In this method~\cite{lee2013pseudo}, an initial set of predictions are estimated for unlabeled images from a network trained with limited annotated labeled set.
Later, the network is trained in an iterative fashion using both labeled images annotations and unlabeled images predictions (pseudo-labels) as proxy ground truths. 
The pseudo-labels estimates are updated after every few epochs of training and we expect their quality to improve through the training.
This have shown improvements for medical image segmentation~\cite{bai2017semi,fan2020inf}. But it has been found~\cite{chapelle2009semi} that if the initial pseudo-label estimates are erroneous, then using them directly in the segmentation loss function can lead to possible degradation of performance.
To avoid this, some methods~\cite{nair2020exploring,yu2019uncertainty,graham2019mild,jungo2019assessing,cao2020uncertainty,mehrtash2020confidence,camarasa2020quantitative} integrate uncertainty or confidence estimates~\cite{blundell2015weight,gal2016dropout,kendall2017uncertainties} of pseudo-labels into self-training to control the quality of pseudo-labels used for training and thereby reduce the negative effects of poor quality of pseudo-labels.

Another method that uses this idea is Noisy Student~\cite{xie2020self}, where two separate models called Teacher-Student networks are used. Teacher model is used to estimate pseudo-labels while student model is trained with ground-truths of labeled set and pseudo-labels of unlabeled set with consistency loss applied over random augmentations, dropout and stochastic depth. Here, the teacher model is replaced with the latest student model after pre-defined number of iterations to estimate refined pseudo-labels.

(b) Consistency regularization:
Such methods~\cite{sajjadi2016regularization,laine2016temporal,tarvainen2017mean} rely on the presumption that different perturbations/data augmented versions of the same image should yield same output label. 
It is expected that the network retains consistency in the output irrespective of changes made to the same image.
The output label distribution loss is minimized with either mean square error or KL divergence loss between the perturbed/augmented samples generated. 
Some popular methods that apply this include pi-model~\cite{sajjadi2016regularization}, temporal ensembling~\cite{laine2016temporal}, mean-teacher~\cite{tarvainen2017mean}, virtual adversarial training~\cite{miyato2018virtual}. 
Following works~\cite{bortsova2019semi,cui2019semi,yu2019uncertainty,li2020transformation,zhou2020deep,fotedar2020extreme,peng2020mutual,fang2020dmnet} apply some variants and combinations of the above ideas for medical image segmentation.

(c) Other methods include adversarial training with GANs~\cite{zhang2017deep,nie2018asdnet,zhang2018translating,zheng2019semi,han2020semi,li2020shape, valvano2021learning}, entropy minimization~\cite{grandvalet2005semi}, and hybrid methods~\cite{berthelot2019mixmatch,berthelot2019remixmatch,sohn2020fixmatch} which use a combination of above methods along with additional regularization terms.

3. Data Augmentation
Prior works have shown that using affine data augmentations~\cite{cirecsan2011high} such as crop, scale, rotation, flip and other augmentations such as random elastic deformations~\cite{ronneberger2015u}, contrast/brightness based intensity augmentations~\cite{hong2017convolutional,perez2018data} improves the baseline for medical imaging tasks.
Also, MixUp augmentations~\cite{zhang2017mixup,eaton2018improving} has shown to yield benefits on medical datasets.
All the above stated methods only work with labeled examples and do not leverage unlabeled images.
Other works have used generative adversarial networks (GANs) to generate additional synthetic image-label pairs~\cite{goodfellow2014generative,costa2018end,shin2018medical,bowles2018gan,cai2019towards,yu2019ea,qin2019pulmonary} that can be later used for training the network along with labeled examples. But these methods require a moderate number of labeled examples to train a stable GAN model. A recent work~\cite{valvano2021learning} trains a stable GAN model with limited annotations of scribbles along with unlabeled data to learn shape priors to obtain high segmentation performance.
However, there are some works that use unlabeled examples along with few labeled examples to train either CNN~\cite{zhao2019data} via image registration or conditional GAN model~\cite{kcipmi2019} optimized to generate synthetic image-label pairs useful for downstream tasks. These methods have shown to perform better than standard augmentations methods.

\section{Methods}

\begin{figure*}[!ht]
    \centering
    \includegraphics[width=1.0\textwidth]{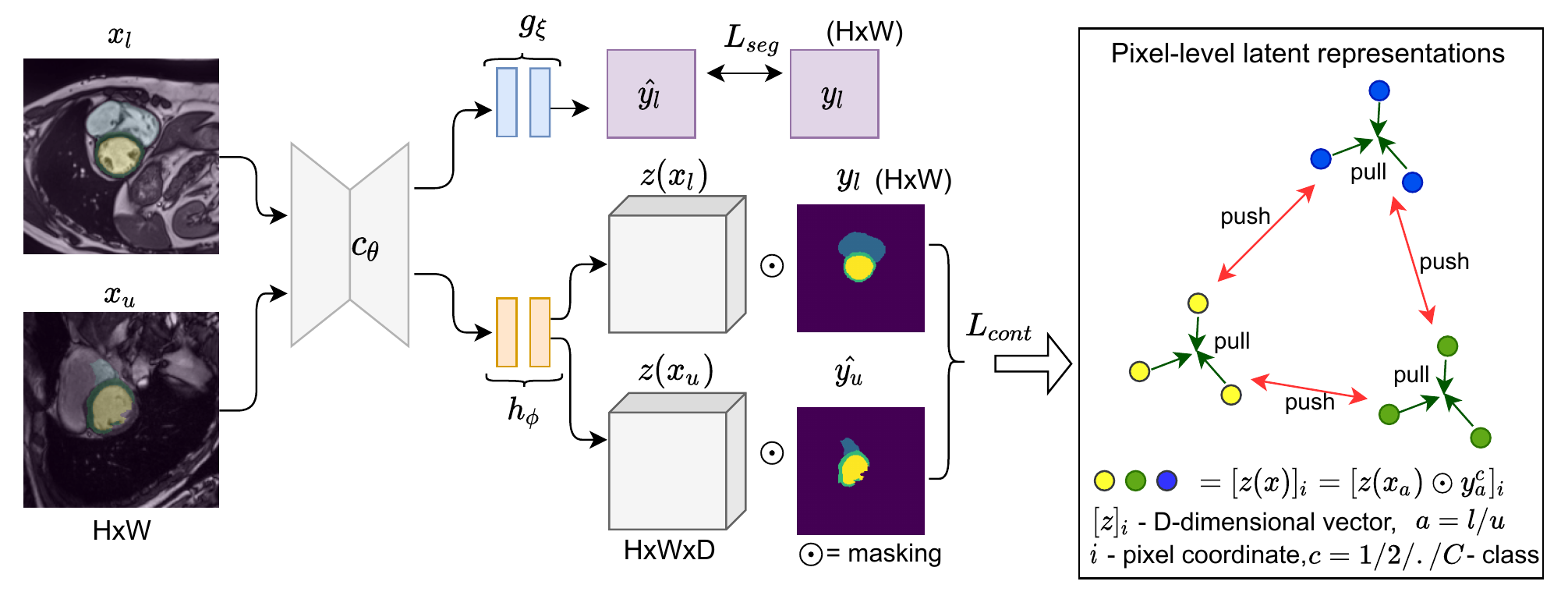}
    \caption{The figure presents the proposed semi-supervised method to get high segmentation performance by applying a pixel-level contrastive loss on pseudo-labels of unlabeled data and limited labeled data along with segmentation loss applied only on labeled set. The pixel-level representations of same class are optimized to be similar and simultaneously optimized to be dissimilar to other class representations in the whole dataset (shown on the right side).}
    \label{fig:methods}
\end{figure*}



The objective of this work is to learn discriminative pixel-level representations with intra-class affinity and inter-class separability by leveraging semantic label information that benefits the semantic segmentation task.
With this objective, in this section, we describe the proposed semi-supervised method in a limited annotation setting.

We first define the following notation that we use for introducing the proposed method.
We define the labeled set with $X_L$ and its image-label pairs with ($x_l, y_l$). Note that in this paper, we focus on the setting where $X_L$ is limited, e.g. $|X_L|=1, 2, 8$. Similarly, $X_U$ is the set of pairs ($x_u, \hat{y}_u$) where $x_u$ denotes unlabeled images and $\hat{y}_u$ denotes the corresponding pseudo-labels obtained and updated using network parameters at certain iterations. The whole dataset is defined as $X = X_L \cup X_U$.

In Figure~\ref{fig:methods}, we present the network architecture that we use in the proposed method where $c_\theta$ is a backbone encoder-decoder network that has two branches: 1) $g_\xi$ is a small network specific for segmentation, 2) $h_\phi$ is another small network specific for training with contrastive loss.

The proposed semi-supervised segmentation algorithm consists of two optimization steps:

\noindent\textbf{In the first optimization step}, we train the networks $c_\theta$ and $g_\xi$ by minimizing a supervised segmentation loss $L_{seg}$ using the limited labeled set $X_L$
\begin{equation}
\begin{split}
\hat{\theta}^{(0)}, \hat{\xi}^{(0)} = \argmin_{\theta, \xi} \frac{1}{|X_L|}\sum_{(x_l, y_l) \in X_L} L_{seg}(y_l, \hat{y}_l)
\end{split}
\label{eq:optimization1}
\end{equation}
where $\hat{y}_l = g_\xi(c_\theta(x_l))$, we use the dice loss proposed by \cite{milletari2016v} for $L_{seg}$ and ($\theta$, $\xi$) are initialized randomly. The purpose of this initial optimization is to obtain reasonable initial pseudo-labels for unlabeled images.

\noindent\textbf{In the second optimization step}, we perform joint training of the networks $c_\theta$, $g_\xi$, and $h_\phi$ using both supervised segmentation loss $L_{seg}$ on $X_L$ and the proposed local contrastive loss $L_{cont}$ on the whole data $X$ that includes both labeled $X_L$ and unlabeled $X_U$ images.

\begin{equation}
\begin{split}
\hat{\theta}, \hat{\xi}, \hat{\phi} = \argmin_{\theta, \xi, \phi} &\frac{1}{|X_L|}\sum_{(x_l, y_l) \in X_L} L_{seg}(y_l, \hat{y}_l) \\&+ \lambda_{cont} \frac{1}{|X|}\sum_{(x, y), (x', y') \in X} L_{cont}((x, y), (x', y')).
\end{split}
\label{eq:optimization2}
\end{equation}
Note that, $(x, y)$ and $(x', y')$ in $L_{cont}$ are two random samples with replacement from $X = X_L \cup X_U$ which can come from either $X_L$ or $X_U$. 
The pseudo-labels of the unlabeled images $x_u \in X_U$ are obtained as $\hat{y}_u = g_{\hat{\xi}^{(t)}}(c_{\hat{\theta}^{(t)}}(x_u))$ using the model parameters $\hat{\xi}^{(t)}$ and $\hat{\theta}^{(t)}$ at iteration $t$.
We initially estimate the pseudo-labels at $t=0$ and then update them in every $P$ iterations. It is also referred to as \textit{pseudo-labeling} step (refer to Section 4.4 and 5.c for more details and analysis).
Also, note that in the second step of optimization, $\theta$ and $\xi$ are initialized as $\hat{\theta}^{(0)}$ and $\hat{\xi}^{(0)}$, respectively and $\phi$ is initialized randomly. $\lambda_{cont}$ is a weighting parameter for the contrastive loss. 
Next, \textbf{we introduce $L_{cont}$, the proposed local pixel-wise contrastive loss}.
In the proposed local pixel-wise contrastive loss, we aim to learn discriminative pixel representations based on their class labels and pseudo-labels.
To achieve this, we optimize to match the pixel representations from the same class label within the image and across different images to be similar while they being dissimilar to the pixel representations that belong to other classes.
After the optimization, pixel representations are expected to form a cluster per class as illustrated on the right part of the Figure \ref{fig:methods}. This allows us to have a better class separation in the feature space by leveraging the contrastive loss on the whole dataset using both ground truth labels of the labeled set and pseudo-label estimates of the unlabeled set.

The feature map after passing an image $x$ through the common network ($c_\theta$) and contrastive branch ($h_\phi$) is denoted by $z(x) = h_{\phi}(c_\theta(x))$ which has dimensions of $H \times W \times D$ where $H$, $W$ are same as input image dimensions and $D$ is the number of channels.
Let us also denote the set of pixel coordinates that belong to a foreground class $c$ for image $x$ as $S_c(x)$ where $1 \leq c \leq C$ and $C$ is the total number of classes to be segmented.
We define the total local contrastive loss between two random samples ($x, y$) and ($x', y'$) from $X$ as
\begin{equation}
\begin{split}
    &L_{cont}((x, y), (x', y')) = \\& \frac{1}{C} \sum_{c=1}^{C} 
    \frac{1}{|S_c(x)|}
    \sum_{i \in S_c(x)} L_{i,c}([z(x)]_i, \bar{z}_c(x'))
    \label{eq:net_local_loss}
\end{split}
\end{equation}
where $[z(x)]_i$ is the feature vector of $x$ at pixel coordinate $i$, which is $D$ dimensional, and 
\[
\bar{z}_c(x') = \frac{1}{|S_c(x')|} \sum_{i \in S_c(x')} [z(x')]_i 
\]
is the mean pixel representation of $x'$ for class $c$. We define the local contrastive loss between a pixel representation feature vector and a mean pixel representation, possibly coming from a different image, $L_{i,c}(., .)$ for a class $c$ as follows:
\begin{equation}
\begin{split}
&L_{i,c}([z(x)]_i, \bar{z}_c(x'))
    = \\& - \log \frac{e^{\mathrm{sim}({[z(x)]_i},{\bar{z}_c(x')})/\tau}}
    {e^{\mathrm{sim}({[z(x)]_i},{\bar{z}_c(x')})/\tau} + \sum_{k \neq c} e^{\mathrm{sim}({[z(x)]_i},{\bar{z}_k(x')})/\tau}}
    \label{eq:local_loss_per_pair}
\end{split}
\end{equation}
where $\mathrm{sim}(a,b) = {a}^{T} b/\|a\|\|b\|$ is the cosine similarity to measure the similarity between two representation vectors, and $\tau$ denotes the temperature scaling factor as defined in~\cite{chen2020simple}.

The proposed local pixel-wise contrastive loss defined in Eq. (\ref{eq:net_local_loss}) could also be designed to match pixel representation ($[z(x)]_i$) of image $x$ at pixel location $i$ to the pixel representation ($[z(x')]_j$) of image $x'$ at location $j$. In a similar way, this would encourage similar representations for pixels belonging to the same class $c$ while simultaneously pushing $[z(x)]_i$ away from $[z(x')]_k$ if the corresponding pixels belong to different classes. For a stable and computationally more efficient training, we match $[z(x)]_i$ to the mean representation vector $\bar{z}_c(x')$ and push it away from the mean representation vectors of other classes $\bar{z}_k(x')$.


Two different implementation of the proposed local loss is possible: 1) matching \textbf{intra-image} pixel representations: When $(x, y) = (x', y')$ in Eq. (\ref{eq:net_local_loss}), the similar class mean representation $\bar{z}_c(x)$ and dissimilar class mean representations $\bar{z}_k(x)$ ($k \neq c$) are computed from the same image.
2) matching \textbf{inter-image} pixel representations: When $(x, y) \neq (x', y')$ in Eq. (\ref{eq:net_local_loss}), $\bar{z}_c(x)$ and $\bar{z}_k(x)$ ($k \neq c$) are computed from different images sampled from $X$.
In the intra-level representation matching, both the similar and dissimilar classes mean representations are computed from  within the same image.
In the inter-level representation matching, for similar pairs, the mean representation of a class is computed from within the same image as well as from different images in the batch. Similarly, the dissimilar mean representations are computed from within the same image and different images in the batch.
With inter-level matching, one can learn more robust representations for the same class structures present in the dataset.


Due to memory limitations, we cannot apply Eq. (\ref{eq:net_local_loss}) to pixel representations for all the coordinates $i \in S_c(x)$ in an image since $|S_c(x)|$ can be very large. 
We deal with this problem by subsampling a smaller set of pixel coordinates $\tilde{S}_c(x) \subset S_c(x)$ in each iteration to fit the GPU memory. 
Then, we use $\tilde{S}_c(x)$ instead of $S_c(x)$ in Eq. (\ref{eq:net_local_loss}).
Also, note that the optimization in Eq. (\ref{eq:optimization2}) can be performed using mini-batch gradient descent by selecting a mini-batch of $X_B$ from $X$ such that $X_B$ contains samples from both $X_L$ and $X_U$.

\section{Experiments}

\subsection{Datasets}
We evaluated the proposed approach on three public MRI datasets.

\noindent\textbf{(a) ACDC Dataset:} It contains 100 short-axis MR-cine T1 3D volumes of cardiac anatomy acquired using 1.5T and 3T scanners. The expert annotations are provided for three structures: right ventricle, myocardium, and left ventricle. It was hosted as part of the MICCAI ACDC challenge 2017.

\noindent\textbf{(b) Prostate dataset:} It contains 48 T2-weighted MRI 3D volumes of prostate. The expert annotations are provided for two structures of prostate: peripheral zone and central gland. 
It was hosted in the medical decathlon challenge in MICCAI 2019.

\noindent\textbf{(b) MMWHS dataset:} It contains 20 T1 MRI 3D volumes of cardiac anatomy with expert annotations for seven structures: left ventricle, left atrium, myocardium, ascending aorta, pulmonary artery, right ventricle, and right atrium.
It was hosted in STACOM and MICCAI 2017 challenges.

\subsection{Pre-processing}
All the images are bias-corrected using N4~\cite{tustison2010n4itk} bias correction using the implementation in the ITK toolkit.
After this, we apply the below pre-processing steps for all the datasets:
(i) each 3D volume ($x$) is normalized using min-max normalization: $(x-x_{1})/(x_{99}-x_{1})$ where $x_{1}$ and $x_{99}$ denote the 1st and 99th percentile in $x$.
(ii) Next, we re-sample all the 2D images and their corresponding labels into a fixed in-plane resolution $r_f$ using bilinear and nearest neighbour interpolation, respectively. This is followed by cropping or zero padding to a fixed image dimensions of $d_f$.
The resolution ($r_f$) and image dimensions ($d_f$) for the datasets are: (a) ACDC: $r_f$ = $1.367 \times 1.367 mm^2$ and $d_f$ = $192 \times 192$, and (b) Prostate: $r_f$ = $0.6 \times 0.6 mm^2$ and $d_f$ = $192 \times 192$, and (c) MMWHS: $r_f$ = $1.5 \times 1.5 mm^2$ and $d_f$ = $160 \times 160$.

\subsection{Network Architecture}
We use an UNet based architecture that consists of a common encoder and decoder networks denoted by $c_\theta$.
From the last layer of the decoder, we have 2 different branches of smaller networks: one for segmentation ($g_\xi$) and one for contrastive learning ($h_\phi$).
The encoder consists of 6 convolutional blocks, each block consists of two $3 \times 3$ convolutions followed by a $2 \times 2$ max pool layer with a stride of 2.
The decoder consists of 5 convolutional blocks, each block consists of an upsampling layer with a factor of 2, followed by concatenation from a corresponding layer of the encoder via a skip connection, that is followed by two $3 \times 3$ convolutions.
The segmentation task network ($g_\xi$) consists of three $3 \times 3$ convolutions, followed by a softmax layer to output the segmentation mask, used in the dice loss~\cite{milletari2016v}.
The contrastive task network ($h_\phi$) consists of two $1 \times 1$ convolution layers that outputs feature maps $f$ of dimensions $H \times W \times D$ that are used for local contrastive loss computation.
We use $1 \times 1$ convolution to mimic the behaviour of dense layers in the projection head in learning global level representations in the popular contrastive learning work~\cite{chen2020simple}.
All the layers except the last layers of both segmentation task and contrastive loss computation have batch normalization~\cite{batchnorm} and ReLU activation layers.

\subsection{Experimental Setup}
(i) For the proposed method, for initial 5000 iterations, we train with only the segmentation loss using the labeled set $X_L = (x_l,y_l)$. In this step, only the common encoder-decoder network parameters ($\theta$) and segmentation-specific layers parameters ($\xi$) are updated.
(ii) Next, we estimate the pseudo-labels $\hat{y_u} \in Y_U$ of unlabeled set $X_U$ using this trained network.
(iii) After this, we perform the optimization given in Eq. (\ref{eq:optimization2}) for $P$=5000 iteration. Here, all the network parameters are updated: encoder-decoder network ($\theta$), segmentation specific layers ($\xi$), and contrastive loss specific layers ($\phi$).
(iv) We iterate the process of pseudo-label initial estimation/re-estimation (ii) and joint loss training (iii) for a total of 3 pseudo-labeling steps in this work resulting in total no of iterations to be 15000 for all the datasets. This value of 3 was found to be a good solution in the compared self-training work~\cite{bai2017semi}.


\subsection{Training Details}
As we mentioned in Sec. 3, the proposed method has two optimization steps. 
We perform both optimizations using Adam optimizer \cite{adamopti} with a learning rate of $10^{-3}$.
We set the batch size $|X_B| = 20$ in both optimization steps.
During the first step we solve the problem given in Eq.~\ref{eq:optimization1}, i.e., we sample the images in $X_B$ only from the limited labeled set $X_L$ and optimize the supervised segmentation loss.
We run this optimization for 5000 iterations.
During the second step we solve the problem given in Eq.~\ref{eq:optimization2}. At each batch, we sample 10 image-label pairs from $X_L$ and 10 image-pseudo-label pairs from $X_U$.
We run this optimization for 15000 iterations where we initially estimate the pseudo-labels at iteration 0 and update them every $P$=5000 iterations.


For the local contrastive loss, we apply random intensity transformations (contrast plus brightness) on the input image batch used to compute the contrastive loss.
For the segmentation loss, we choose dice loss~\cite{milletari2016v} and apply random data augmentations such as crop, scale, rotation, flip, random elastic deformations, and random intensity transformations of contrast and brightness on the batch of labeled images. 

For the contrastive loss, we sample $|\tilde{S}_c(x)|$ number of positive pixel representations for each class $c$ and each image $x$. The default value of $|\tilde{S}_c(x)|$ is set as 3 for all the experiments unless stated otherwise. The number of classes $C$ (structures of interest) is different for each dataset, whose value is 3 for ACDC, 2 for Prostate, and 7 for MMWHS.
The temperature parameter $\tau$ chosen for contrastive loss is 0.1 as defined in~\cite{chaitanya2020contrastive} for the medical datasets.
We use the number of channels as 16 in the last layer of the contrastive network branch $h_\phi$. 
Therefore, the pixel representations are 16-dimensional for all the datasets.
We used the highest Dice score model on the validation set during training to select the model for the final evaluation on the test set.

\textbf{Dataset Split:} We use the following size of unlabeled ($X_U$) and test sets ($X_{ts}$): (a) $|X_U|$=52, $|X_{ts}|$=20 for ACDC, (b) $|X_U|$=22, $|X_{ts}|$=15 for Prostate, and (c) $|X_U|$=10, $|X_{ts}|$=10 for MMWHS.
For the limited annotation setting evaluation, we use a small labeled ($X_L$) and validation sets ($X_{val}$). The validation set ($|X_{val}|$) size is always fixed to two 3D volumes for all datasets.
For the labeled set, we experiment with three different sizes of $|X_{L}|$ = 1, 2, and 8 3D volumes. 

\textbf{Evaluation}:
The segmentation performance is evaluated using Dice's similarity coefficient (DSC)~\cite{dice1945measures}. For all the experiments, we report the mean DSC over all the structures, without background, in the test set $|X_{ts}|$ over 6 different runs.
For each run, we construct different sets of $X_L$ and $X_{val}$ by randomly sampling from the available set of volumes.


\section{Results and Discussion}

\begin{table*}[!t]
\begin{center}
\begin{tabular}{|p{1.8cm}|p{1.0cm}|p{1.0cm}|p{1.0cm}|p{1.0cm}|p{1.0cm}|p{1.0cm}|p{1.0cm}|p{1.0cm}|p{1.0cm}|p{1.0cm}|}
\hline
\rowcolor{lightergray} \multicolumn{2}{|c|}{} & \multicolumn{3}{c|}{ACDC} & \multicolumn{3}{c|}{Prostate} & \multicolumn{3}{c|}{MMWHS} \\ \cline{3-11} 
\rowcolor{lightergray} \multicolumn{2}{|c|}{Method} & {$|X_{L}|$=1} & {$|X_{L}|$=2} & {$|X_{L}|$=8} & {$|X_{L}|$=1} & {$|X_{L}|$=2} & {$|X_{L}|$=8} & {$|X_{L}|$=1} & {$|X_{L}|$=2} & {$|X_{L}|$=8} \\ \cline{1-11}
\rowcolor{lightgray} \multicolumn{11}{|c|}{{{Baseline}}} \\ 
\rowcolor{lightergray} \multicolumn{2}{|c|}{Random init.} & 0.614 & 0.702 & 0.844 & 0.489 & 0.550 & 0.636 & 0.451 & 0.637 & 0.787   \\ 
\rowcolor{lightgray} \multicolumn{11}{|c|}{{{Semi-supervised Methods}}} \\ 
\rowcolor{lightergray} \multicolumn{2}{|c|}{Self-training~\cite{bai2017semi}} & 0.690 & 0.749 & 0.860 & 0.551 & 0.598 & 0.680 & 0.563 & 0.691 & 0.801 \\ 
\rowcolor{lightergray} \multicolumn{2}{|c|}{Mixup~\cite{zhang2017mixup}} & 0.695 & 0.785 & 0.863 & 0.543 & 0.593 & 0.661 & 0.561 & 0.690 & 0.796 \\ 
\rowcolor{lightergray} \multicolumn{2}{|c|}{Data Augment~\cite{kcipmi2019}} & 0.731 & 0.786 & 0.865 & 0.585 & 0.597 & 0.667 & 0.529 & 0.661 & 0.785 \\ 
\rowcolor{lightergray} \multicolumn{2}{|c|}{Adversarial Tr.~\cite{zhang2017deep}} & 0.536 & 0.654 & 0.791 & 0.487 & 0.544 & 0.586 & 0.482 & 0.655 & 0.779 \\ 
\rowcolor{lightergray} \multicolumn{2}{|c|}{Noisy Student~\cite{xie2020self}} & 0.632 & 0.737 & 0.836 & 0.556 & 0.601 & 0.668 & 0.593 & 0.685 & 0.780 \\ 
\rowcolor{lightgray} \multicolumn{11}{|c|}{{{Proposed Method}}} \\ \rowcolor{lightergray} \multicolumn{2}{|c|}{Pseudo-labels joint Tr.  (intra)} & \underline{0.761} & \underline{0.845} & 0.881 & 0.571 & 0.613 & 0.693 & \underline{0.599} & \underline{0.721} & 0.803 \\ 
\rowcolor{lightergray} \multicolumn{2}{|c|}{Pseudo-labels joint Tr.  (inter)} & 0.759 & 0.831 & \underline{0.883} & \underline{0.578} & \underline{0.618} & \underline{0.696} & 0.572 & 0.719 & \underline{0.811} \\ 
\rowcolor{lightgray} \multicolumn{11}{|c|}{{{Benchmark}}} \\ 
\rowcolor{lightergray} \multicolumn{2}{|c|}{Training with large $|X_{L}|$} & \multicolumn{3}{|r|}{($|X_{L}| = 78$) 0.912} & \multicolumn{3}{|r|}{($|X_{L}| = 20$) 0.697} & \multicolumn{3}{|r|}{($|X_{L}| = 8$) 0.787}\\ \hline

\end{tabular}
\caption{Comparison of the proposed method with other semi-supervised learning, data augmentation methods and concurrent contrastive learning works. The proposed pseudo-label joint training provides better results than compared methods for all datasets and most $|X_{tr}|$ values. 
In each column, best values among all methods are underlined.}
\label{table:main_table_comparison}
\end{center}
\end{table*}

In Table~\ref{table:main_table_comparison}, we present the results of baseline, existing semi-supervised methods in the literature, the proposed method, and concurrent contrastive learning works.

\subsection{Baseline and semi-supervised methods}
From Table~\ref{table:main_table_comparison}, we can observe that semi-supervised methods yield a significant boost in performance for limited labeled settings of $|X_{L}|$=1 or 2 3D volumes for all the datasets compared to the baseline.
We observe that methods like self-training and noisy student with $|X_{L}|$=2 close the gap to benchmark that is trained with a large number of annotations to less than 0.1 DSC for Prostate and MMWHS datasets, and 0.2 DSC for ACDC dataset.

\subsection{Proposed method}
We observe that the proposed method yields better results than the baseline and compared semi-supervised methods.
Results suggest that using pseudo-labels only in a contrastive loss, as done in the proposed method, yields higher performance gains compared to using them in the segmentation loss, as is often done in popular self-training methods in medical image analysis. 
This can be due to initial pseudo-label estimates being erroneous and thus deteriorating the performance if directly used for the segmentation loss optimization, especially for the limited labeled case of $|X_{L}|$=1 or 2 3D volumes.
We hypothesize that using pseudo-labels only in the contrastive loss and not in the segmentation loss acts as regularization, and prevent erroneous pseudo-label mask information propagating into the segmentation task-specific layers.
We believe that this provides us with the additional improvements that could have hindered the self-training approach. Also, another reason for improvements can be due to learning a stronger pixel-level representation by matching representations of the same class to be similar across images with the defined contrastive loss, which typical segmentation loss of cross-entropy or dice loss cannot learn.
Also, we see that the re-estimation of pseudo-labels for the unlabeled data iteratively in the training leads to continuous improvements through the training as as shown in Figure~\ref{fig:x_tst_dsc_self_train_vs_ours} (more discussion in section 5.5). 

We evaluated two pixel representation matching schemes: intra-image and inter-image pixel representation matching. Both matching schemes yield improvements over compared methods.
Inter-image pixel representation matching performs better than intra-image pixel representation matching in 5 out of 9 cases. This can be because matching pixel representations across images from different volumes from the same class can provide the network additional cues over intra-image matching to learn more robust representations for each semantic class in the dataset.
For a limited labeled case of $X_{L}$=1, the intra-image matching performs better on ACDC and MMWHS datasets than inter-image as the quality of the pseudo-labels can be of low quality in this case, and thereby, matching erroneous class pixel representations across images can hinder the performance gains that can be obtained. 

\subsection{Concurrent contrastive semi-supervised methods}
In the concurrent works~\cite{alonso2021semi,zhao2020contrastive}, a different set of semi-supervised contrastive learning frameworks are proposed. 
In~\cite{zhao2020contrastive}, a two step training is devised of pre-training with contrastive loss and later fine-tuning to final task with labeled set, which is a fully supervised setting.
The two step training strategy can also be used in semi-supervised setting by using both the labeled annotations and pseudo-labels of the unlabeled set for both stages of the training.
We perform experiments in both settings and present these results in the first two rows of Table~\ref{table:concurrent_cont_lr_works} in fully supervised and semi-supervised settings, respectively.
With the results presented, we can infer that joint training can be more beneficial than disjoint training, and using pseudo-labels in segmentation loss can hinder the maximum possible gains that could be obtained, especially, in a limited labeled setting of $|X_{L}|$= 1 and 2.
For~\cite{alonso2021semi}, they perform joint training where both contrastive and segmentation losses are computed using both labeled and unlabeled sets. They devise student-teacher networks, use additional memory bank to store class representations and use entropy minimization. 
We rather propose a simpler solution without using an additional teacher network, memory banks and entropy minimization.
For an equivalent comparison with the proposed approach, we evaluate this joint training approach with only one network, and without additional components. Here, both the labeled annotations and pseudo-labels of the unlabeled set are used in both the contrastive and segmentation losses. 
We present these results in third row of table~\ref{table:concurrent_cont_lr_works}, where we see that using pseudo-labels in segmentation loss computation even in joint training can lead to lower gains similar to our observation in the self-training approach.
Overall, we obtain higher gains than these methods for all settings across all the datasets where the gains are more prominent for the ACDC dataset.  

\begin{table*}[!t]
\begin{center}
\begin{tabular}{|p{1.8cm}|p{1.0cm}|p{1.0cm}|p{1.0cm}|p{1.0cm}|p{1.0cm}|p{1.0cm}|p{1.0cm}|p{1.0cm}|p{1.0cm}|p{1.0cm}|}
\hline
\rowcolor{lightergray} \multicolumn{2}{|c|}{} & \multicolumn{3}{c|}{ACDC} & \multicolumn{3}{c|}{Prostate} & \multicolumn{3}{c|}{MMWHS} \\ \cline{3-11} 
\rowcolor{lightergray} \multicolumn{2}{|c|}{Method} & {$|X_{L}|$=1} & {$|X_{L}|$=2} & {$|X_{L}|$=8} & {$|X_{L}|$=1} & {$|X_{L}|$=2} & {$|X_{L}|$=8} & {$|X_{L}|$=1} & {$|X_{L}|$=2} & {$|X_{L}|$=8} \\ \cline{1-11}
\rowcolor{lightergray} \multicolumn{2}{|c|}{Two step Tr.~\cite{zhao2020contrastive} (fully-sup)} & 0.568 & 0.717 & 0.841 & 0.561 & 0.596 & 0.686 & 0.509 & 0.661 & 0.795 \\
\rowcolor{lightergray} \multicolumn{2}{|c|}{Two step Tr.~\cite{zhao2020contrastive} (semi-sup)} & 0.622 & 0.747 & 0.843 & 0.568 & 0.606 & 0.688 & 0.559 & 0.704 & 0.804 \\
\rowcolor{lightergray} \multicolumn{2}{|c|}{Joint Tr. with pseudo-labels in seg. loss} & 0.420 & 0.603 & 0.749 & 0.435 & 0.495 & 0.545 & 0.470 & 0.510 & 0.638 \\ \hline
\rowcolor{lightergray} \multicolumn{2}{|c|}{Proposed joint Training (inter)} & \underline{0.759} & \underline{0.831} & \underline{0.883} & \underline{0.578} & \underline{0.618} & \underline{0.696} & \underline{0.572} & \underline{0.719} & \underline{0.811} \\ 
\hline 
\end{tabular}
\caption{We compare with some concurrent contrastive learning works. We observe proposed approach to give better results with the improvements being higher for ACDC dataset while them being relatively smaller for Prostate and MMWHS datasets.}
\label{table:concurrent_cont_lr_works}
\end{center}
\end{table*}

\subsection{Better model initialization from contrastive loss pre-training}
Recent work by~\cite{chaitanya2020contrastive} applies a contrastive learning framework to learn a good initialization for medical image segmentation tasks and achieve state-of-the-art results and perform better than random initialization.
We use this pre-trained network initialization for training the baseline, self-training, and proposed method whose results are presented in Table~\ref{table:prop_init_comparison}.
Firstly, we obtain larger improvements when fine-tuning a baseline model with this pre-trained initialization over random initialization.
Secondly, we get higher mean DSC for both self-training and the proposed method with this initialization compared to the random initialization. This is because we have a better initial baseline model that yields more accurate initial pseudo-label estimates compared to the random initialization.
Thirdly, for this initialization as well, we get higher gains with the proposed method over self-training where the gains on the ACDC dataset are more significant while the gains are smaller on the remaining two datasets.
Lastly, we see that proposed joint training is complementary to a good network initialization learned via pre-training.


\begin{table*}[!t]
\begin{center}
\begin{tabular}{|p{1.8cm}|p{1.0cm}|p{1.0cm}|p{1.0cm}|p{1.0cm}|p{1.0cm}|p{1.0cm}|p{1.0cm}|p{1.0cm}|p{1.0cm}|p{1.0cm}|}
\hline
\rowcolor{lightergray} \multicolumn{2}{|c|}{} & \multicolumn{3}{c|}{ACDC} & \multicolumn{3}{c|}{Prostate} & \multicolumn{3}{c|}{MMWHS} \\ \cline{3-11} 
\rowcolor{lightergray} \multicolumn{2}{|c|}{Method} & {$|X_{L}|$=1} & {$|X_{L}|$=2} & {$|X_{L}|$=8} & {$|X_{L}|$=1} & {$|X_{L}|$=2} & {$|X_{L}|$=8} & {$|X_{L}|$=1} & {$|X_{L}|$=2} & {$|X_{L}|$=8} \\ \cline{1-11}
\rowcolor{lightergray} \multicolumn{2}{|c|}{PI. + Baseline} & 0.725 & 0.789 & 0.872 & 0.579 & 0.619 & 0.684 & 0.569 & 0.694 & 0.794 \\ 
\rowcolor{lightergray} \multicolumn{2}{|c|}{PI. + Self-training(~\cite{bai2017semi})} & 0.745 & 0.802 & 0.881 & 0.607 & 0.634 & \underline{0.698} & 0.647 & 0.727 & 0.806 \\ 
\rowcolor{lightergray} \multicolumn{2}{|c|}{PI. + Proposed method (inter)} & \underline{0.774} & \underline{0.845} & \underline{0.887} & \underline{0.612} & \underline{0.641} & 0.692 & \underline{0.651} & \underline{0.734} & \underline{0.814} \\ \hline 
\end{tabular}
\caption{We compare baseline, self-training, and proposed method trained over a pre-trained network initialization (PI.) instead of random initialization. This initialization was learnt via contrastive loss based pre-training as done in~\cite{chaitanya2020contrastive}. We see that both self-training and proposed methods benefit from better initialization and lead to higher gains with proposed approach yielding better improvements.}
\label{table:prop_init_comparison}
\end{center}
\end{table*}

\subsection{Visualization of results of the proposed method}
\begin{figure}[!t]
    \centering
    \includegraphics[width=1.05\columnwidth]{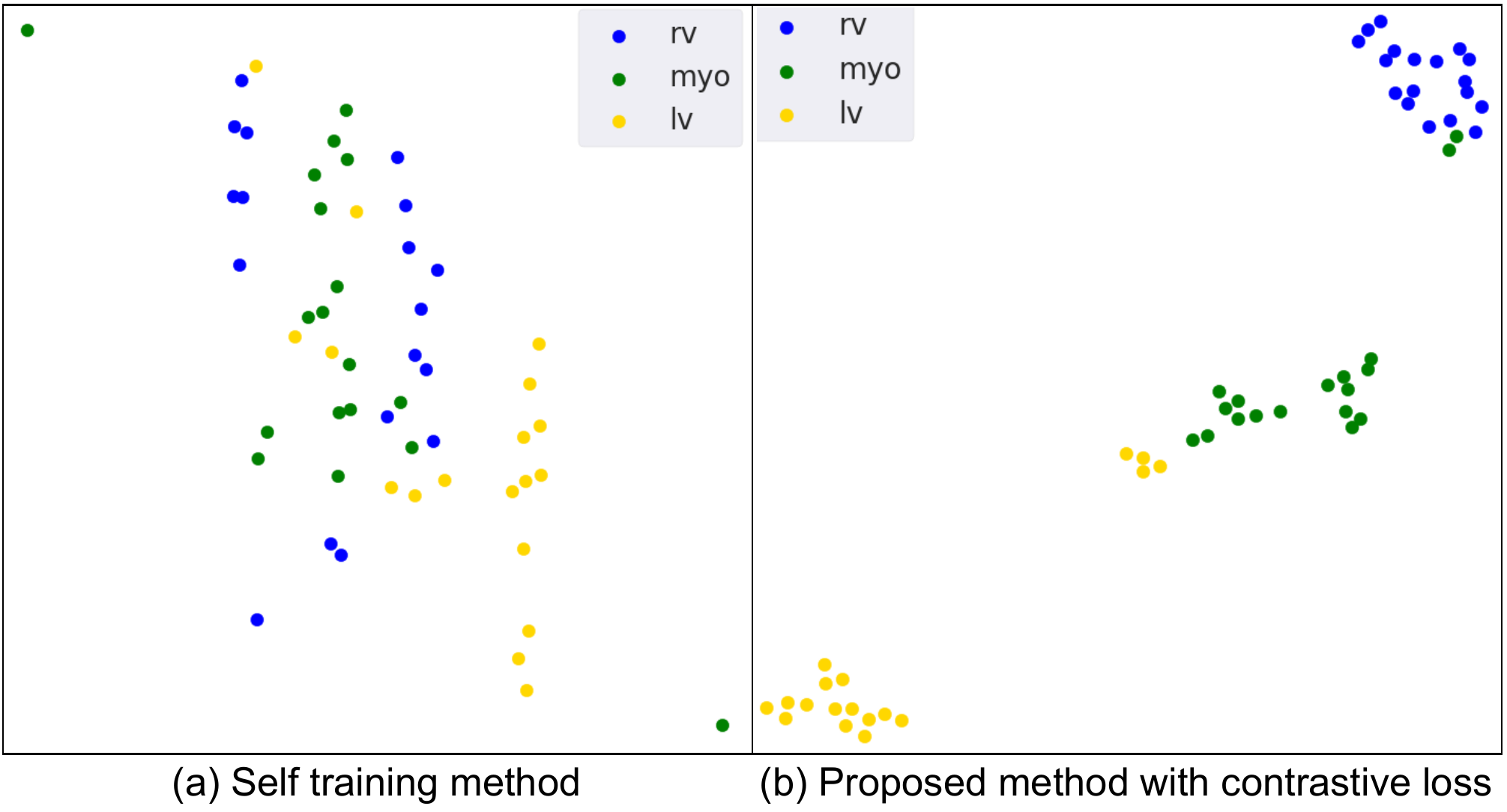}
    \caption{tSNE plot of pixel representations of the last layer of the common network ($c_\theta$ for three structures of ACDC (right ventricle (rv), myocardium (myo), left ventricle (lv)) for some randomly sampled unlabeled images for the methods: (a) self-training and (b) proposed method with local contrastive loss. For the proposed approach, we observe a better intra-class affinity and inter-class seperation between pixel representations compared to self-training, and thereby leading to higher segmentation performance.}
    \label{fig:tsne_plot}
\end{figure}

In Figure~\ref{fig:tsne_plot}, we present a tSNE plot of the pixel representations for the three cardiac structures of the right ventricle (rv), myocardium (myo), left ventricle (lv) of the ACDC dataset at the last layer of the common network (output of $c_{\theta}$) for both the self-training model and proposed work model.
In the proposed method, we observe that the pixel representations of the three structures have more compact clusters for each class (intra-class affinity), and different class representations are more separable (inter-class separability) compared to the pixel representations from the self-training model. With the contrastive loss, the network learns a better intra-class affinity and inter-class separability compared to a segmentation loss that may not be optimal to learn such semantic relationships between pixels.

\begin{figure}[!t]
    \centering
    \includegraphics[width=1.0\columnwidth]{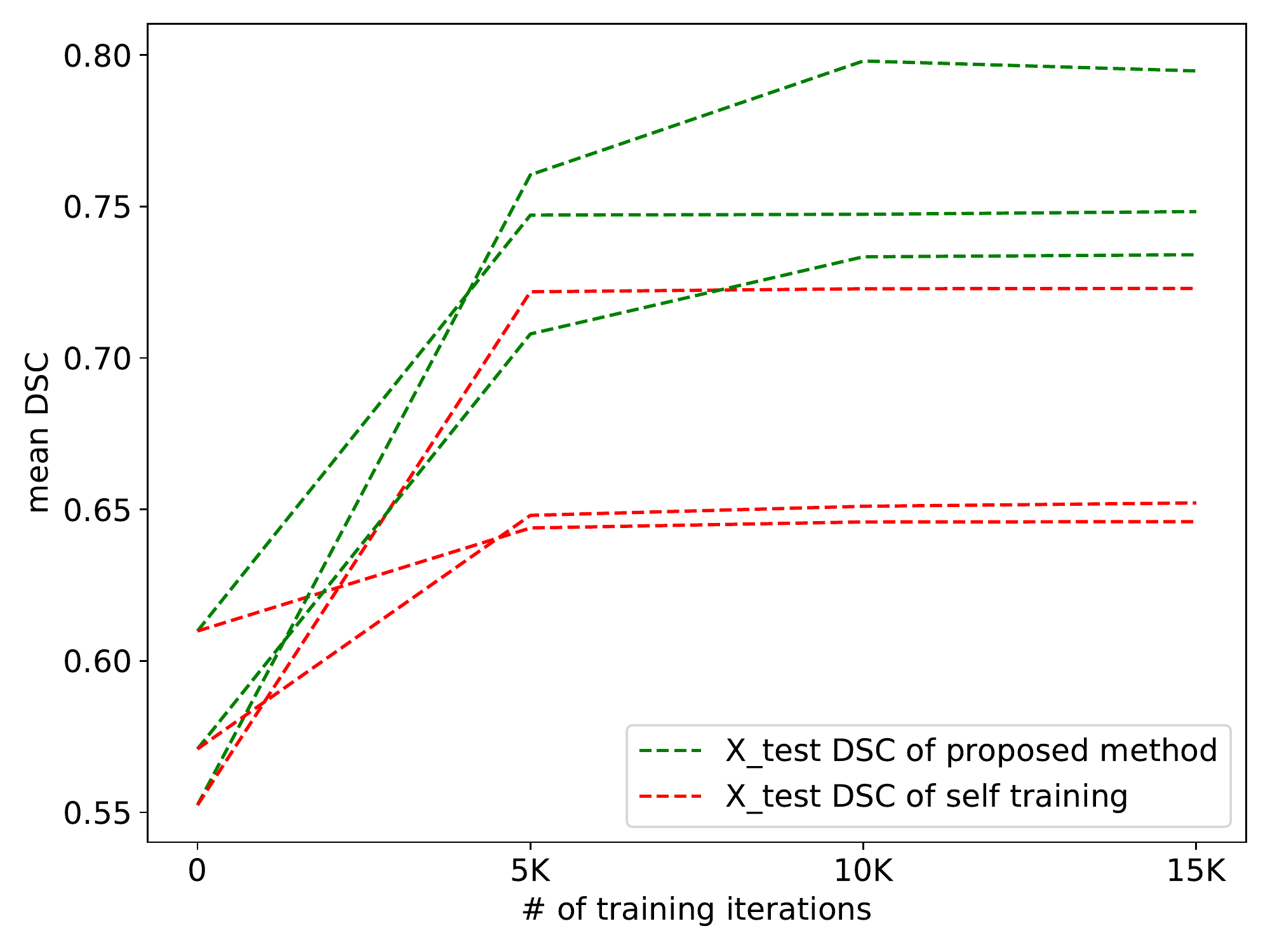}
    \caption{Test set mean DSC reported on ACDC dataset over the 3 pseudo-labeling steps for self-training and proposed method, where each step denotes 5000 iterations of training. The proposed approach yields consistent and higher performance gains for first and second pseudo-labeling steps compared to self-training that do not provide any additional gains from second update step.}
    \label{fig:x_tst_dsc_self_train_vs_ours}
\end{figure}

In Figure~\ref{fig:x_tst_dsc_self_train_vs_ours}, for the setting of $|X_L|$=1 3D volume on ACDC dataset, we present the test set's mean DSC for three runs where a different set of training volume and validation volumes chosen for self-training and proposed method. 
The initial model was trained using supervised segmentation loss using only a limited labeled set of $X_L$=1. Later, we use this model to estimate pseudo-labels such that both methods start with same pseudo-label estimates. 
After this, we perform the respective iterative training as proposed in self-training and proposed approach with three pseudo-labeling steps with pseudo-labels updated every 5000 iterations.
Here, we observe that the test set's mean DSC improves after the three pseudo-labeling steps for both self-training and proposed method. 
The improvements are larger for the proposed method from the first pseudo-labeling step compared to self-training across all runs.
Also, we see that the self-training approach's performance gains become constant from second pseudo-labeling step while the proposed approach yields further improvements at the second step.
After the third update step, even proposed method's gains become constant across all 3 runs.

\subsection{Hyper-parameter Analysis}
Since a large number of computational resources are required for each hyper-parameter experiment on each dataset, we evaluate different hyper-parameters on ACDC dataset only.

\noindent \textbf{(a) $\lambda_{cont}$ for contrastive loss ($L_{cont}$)}: In Table~\ref{tab:lambda_vary}, we present the results for different values of lambda ($\lambda_{cont}$) as defined in Eq.~(\ref{eq:optimization2}) to dictate the contribution of the contrastive loss in the total loss used for updating the network weights. We observe that the performance deteriorates when we use a very small lambda value for the contrastive loss contribution when the number of labeled examples is low. This coefficient does not have a distinguishable effect when $|X_L| = 8$.

\begin{table}[!t]
  \centering
    \begin{tabular}{|p{2.5cm}|p{1.0cm}|p{1.0cm}|p{1.0cm}|p{1.0cm}|p{1.0cm}|}
    \hline
    \rowcolor{lightergray} {representation matching scheme} & $\lambda_{cont}$ value & {$|X_{L}|$=1} & {$|X_{L}|$=2} & {$|X_{L}|$=8} \\
    \hline
    \rowcolor{lightergray} {Intra} & 1 & 0.763 & 0.817 & 0.886 \\ 
    \rowcolor{lightergray} {Intra} & 0.1 & 0.761 & 0.845 & 0.881 \\ 
    \rowcolor{lightergray} {Intra} & 0.01 & 0.731 & 0.801 & 0.890 \\ \hline
    \rowcolor{lightergray} {Inter} & 1 & 0.749 & 0.849 & 0.898 \\  
    \rowcolor{lightergray} {Inter} & 0.1 & 0.759 & 0.831 & 0.883 \\ 
    \rowcolor{lightergray} {Inter} & 0.01 & 0.722 & 0.815 & 0.888 \\ \hline 
  \end{tabular}
  \caption{Fine-tuning with different lambda ($\lambda_{cont}$) values to control the influence of contrastive loss in the proposed method with a random network initialization on ACDC dataset. We observe the proposed method to be relatively stable for $\lambda_{cont}$ values of 0.1, 1 and the performance drops slightly for low value of 0.01.}
  \label{tab:lambda_vary}
\end{table}

\noindent \textbf{(b) Number of pseudo-labeling steps:} 
Here, we analyze the effect of number of pseudo-labeling steps during the training.
We define the total number of training iterations to be 15000 for all the below experiments. We evaluate three values of pseudo-labeling steps of 2, 3, and 4 where the pseudo-labels are updated in every $P$ iterations of $P$=7500 (15000/2), $P$=5000 (15000/3), and $P$=3750 (15000/4), respectively.
The results are presented in Table~\ref{tab:pseudo_label_update_freq}.
We observe for both intra and inter-representation matching schemes that the performance difference between the evaluated values of number of pseudo-labeling steps are small indicating the stable behaviour of training and gains in performance obtained with the proposed approach.

\begin{table}[!t]
  \centering
    \begin{tabular}{|p{2.5cm}|p{2.0cm}|p{1.0cm}|p{1.0cm}|p{1.0cm}|p{1.0cm}|}
    \hline
    \rowcolor{lightergray} {representation matching scheme} & No. of pseudo-labeling steps & {$|X_{L}|$=1} & {$|X_{L}|$=2} & {$|X_{L}|$=8} \\
    \hline
    \rowcolor{lightergray} {Intra} & 2 & 0.764 & 0.836 & 0.878\\ 
    \rowcolor{lightergray} {Intra} & 3 & 0.761 & 0.845 & 0.881 \\ 
    \rowcolor{lightergray} {Intra} & 4 & 0.771 & 0.813 & 0.884 \\ \hline
    \rowcolor{lightergray} {Inter} & 2 & 0.728 & 0.839 & 0.895 \\  
    \rowcolor{lightergray} {Inter} & 3 & 0.759 & 0.831 & 0.883 \\ 
    \rowcolor{lightergray} {Inter} & 4 & 0.777 & 0.83 & 0.888 \\ \hline 
  \end{tabular}
  \caption{We evaluate the effect on segmentation performance when we vary the pseudo-labeling steps on ACDC dataset. We do not observe significant difference between different values of pseudo-labeling steps.}
  \label{tab:pseudo_label_update_freq}
\end{table}

\noindent \textbf{(c) Number of positive pixel representations $|S_c(x)|$ sampled per class $c$ per image $x$:} Here, the number of positive pixel representations $|S_c(x)|$ sampled per class per image to match its class mean representation are varied between the values of 3, 5, and 10.
In Table~\ref{tab:positives_vary}, for both intra- and inter-image representation matching schemes, we do not see a discernible pattern in performance with respect to the number of pixels sampled.
This shows that the model is stable with respect to the final performance against changes in the number of positive pixel representations sampled per class per image.

\begin{table}[!t]
  \centering
    \begin{tabular}{|p{2.5cm}|p{1.0cm}|p{1.0cm}|p{1.0cm}|p{1.0cm}|p{1.0cm}|}
    \hline
    \rowcolor{lightergray} {representation matching scheme} & $|S_c(x)|$ & {$|X_{L}|$=1} & {$|X_{L}|$=2} & {$|X_{L}|$=8} \\
    \hline
    \rowcolor{lightergray} {Intra} & 3 & 0.761 & 0.845 & 0.881 \\ 
    \rowcolor{lightergray} {Intra} & 5 & 0.772 & 0.847 & 0.882 \\ 
    \rowcolor{lightergray} {Intra} & 10 & 0.752 & 0.818 & 0.885 \\ \hline
    \rowcolor{lightergray} {Inter} & 3 & 0.759 & 0.831 & 0.883 \\  
    \rowcolor{lightergray} {Inter} & 5 & 0.750 & 0.847 & 0.882 \\ 
    \rowcolor{lightergray} {Inter} & 10 & 0.732 & 0.862 & 0.897 \\ \hline 
  \end{tabular}
  \caption{Fine-tuning with different number of positive representation examples $|S_c(x)|$ per class $c$ and per image $x$ for the proposed method with a random network initialization on ACDC dataset. We do not see any major differences in performance denoting the stable training.}
  \label{tab:positives_vary}
\end{table}

\noindent \textbf{(d) Improving the quality of the pseudo-labels of unlabeled set:} Here, we control the quality of the pseudo-label predictions of unlabeled volumes used in the contrastive loss, and the results are presented in Table~\ref{tab:pseudo_label_quality}.
The threshold value defined denotes the DSC overlap between two estimated masks of the same volume under two different sets of transformations like rotation, translation, scaling, etc. 
This idea is borrowed from the consistency regularization works~\cite{laine2016temporal,bortsova2019semi,li2020transformation} in the literature to ensure that the pseudo-label predictions of the unlabeled set are accurate and do not vary drastically under transformations. 
The value of $0.9$ denotes having a highly confident estimate where the two estimated masks of the same volume under different transformations are very similar and have an overlap dice of $0.9$ while $0.7$ denotes a less confident estimate where the two estimated masks are less similar to each other.
The value of $0$ denotes that no consistency regularization was applied, and all the pseudo-label predictions of all volumes were used in the contrastive loss.
We observe that using highly confident unlabeled set predictions does not necessarily yield higher gains compared to using all the predictions of the unlabeled set.
At threshold value of 0, we also observe that inter-image representation matching performs worse than intra-image representation matching for lower number of labeled examples of $|X_L|$=1 or 2. This can be due to the poor quality of pseudo-labels estimated, which when used to match the representations of same class across subjects leads to lower gains.

\begin{table}[!t]
  \centering
    \begin{tabular}{|p{2.5cm}|p{1.1cm}|p{1.0cm}|p{1.0cm}|p{1.0cm}|p{1.0cm}|}
    \hline
    \rowcolor{lightergray} {representation matching scheme} & threshold value & {$|X_{L}|$=1} & {$|X_{L}|$=2} & {$|X_{L}|$=8} \\
    \hline
    \rowcolor{lightergray} {Intra} & 0 & 0.761 & 0.845 & 0.881 \\ 
    \rowcolor{lightergray} {Intra} & 0.9 & 0.748 & 0.841 & 0.890 \\ 
    \rowcolor{lightergray} {Intra} & 0.8 & 0.762 & 0.839 & 0.888 \\ 
    \rowcolor{lightergray} {Intra} & 0.7 & 0.769 & 0.856 & 0.878 \\ \hline
    \rowcolor{lightergray} {Inter} & 0 & 0.759 & 0.831 & 0.883 \\  
    \rowcolor{lightergray} {Inter} & 0.9 & 0.779 & 0.832 & 0.891 \\ 
    \rowcolor{lightergray} {Inter} & 0.8 & 0.791 & 0.860 & 0.882 \\ 
    \rowcolor{lightergray} {Inter} & 0.7 & 0.732 & 0.847 & 0.889 \\ \hline 
  \end{tabular}
  \caption{Fine-tuning with different threshold values to control the quality of pseudo-labels of the unlabeled volumes used in the contrastive loss computation of the proposed method with a random network initialization on ACDC dataset. We see that selecting higher quality of pseudo-labels with lower threshold values of 0.7 and 0.8 does not lead to more gains over using all the pseudo-labels (threshold value=0).}
  \label{tab:pseudo_label_quality}
\end{table}

\section{Conclusion}
It is challenging to deploy deep learning-based models with high performance for medical image segmentation due to the requirement of a large set of annotations.
To alleviate this, we propose a semi-supervised method that uses many unlabeled images and a limited set of annotations to yield high segmentation performance.
We introduce a joint training framework where a pixel-wise contrastive loss is defined over pseudo-labels of unlabeled images and limited labeled images with the traditional segmentation loss applied on only the labeled set.
We perform iterative training to improve the quality of the pseudo-labels.
With the proposed contrastive loss, we learn better intra-class compactness and inter-class separability for the segmented classes in the dataset compared to typical segmentation loss-based methods like self-training.
We perform an extensive evaluation of the proposed method on three MRI datasets and obtain high segmentation performance in limited annotation settings. We get higher performance gains than the compared state-of-the-art semi-supervised methods and concurrent contrastive learning methods.
We also show that a good network initialization learned via pre-training with unlabeled images is complementary to the proposed method and can be combined to obtain higher gains. 
Also, using only the labeled set in the segmentation loss compared to earlier approaches is one of the essential detail in the performance gains obtained in the proposed method.

\section{Acknowledgements}

The presented work was partly funding by: 
1. Clinical Research Priority Program (CRPP) Grant on Artificial Intelligence in Oncological Imaging Network, University of Zurich, 
2. Swiss Platform for Advanced Scientific Computing (PASC), coordinated by Swiss National Super-computing Centre (CSCS), 
3. Personalized Health and Related Technologies (PHRT), project number 222, ETH domain, 
4. University Hospital Zurich.
We also thank Nvidia for their GPU donation.

\bibliographystyle{IEEEtran}
\bibliography{references.bib}



\end{document}